\documentclass[sigconf,nonacm]{acmart}

\renewcommand\footnotetextcopyrightpermission[1]{}
\usepackage{listings}
\usepackage{xcolor}
\usepackage{booktabs}
\usepackage{multirow}
\usepackage{balance}
\usepackage{url}
\usepackage[htt]{hyphenat}
\begin{document}

\title{The Energy Blind Spot: NVIDIA's Flagship Edge AI Hardware\\Cannot Support Process-Level Energy Attribution}

\author{Deepak Panigrahy}
\affiliation{
  \institution{Independent Researcher}
  \city{Bentonville}
  \country{USA}
}
\email{deepak.panigrahy03@gmail.com}

\author{Aakash Tyagi}
\affiliation{%
  \institution{Texas A\&M University}
  \city{College Station}
  \state{Texas}
  \country{USA}
}
\email{tyagi@tamu.edu}

\begin{abstract}
Agentic AI workloads — where a single user goal triggers multi-step orchestration, tool calls, retries, 
and failure recovery — are being targeted for edge deployment, with NVIDIA, Dell, HP, ASUS, MSI, Acer, and Gigabyte all 
shipping GB10-based desktop AI systems in 2026~\cite{nvidia2026dgxspark,insiderllm2026gb10}. Prior work shows orchestration 
structure dominates agentic energy cost~\cite{panigrahy2026epg} and CPU-side processing accounts for up to 44\% of total dynamic energy~\cite{rajat2025cpucentric}. 
We report a systematic energy-observability audit of the ASUS Ascent GX10 (GB10 SoC) and find that the platform exposes no CPU energy 
counter, no INA power-rail monitor, no IPMI/BMC, and no SCMI powercap protocol through any supported software interface. The only on-device 
energy telemetry is instantaneous GPU power via NVML. We further discover that the MediaTek firmware already computes per-rail energy 
internally via an undocumented ACPI interface (SPBM), but NVIDIA states there are "no plans to expose CPU rail 
information.''~\cite{nvforum2026cpu}. On-device per-process energy attribution — as performed on x86 via RAPL — is therefore not reproducible on this platform through supported interfaces. We formalize a hardware requirements specification for energy-attributed AI, propose an interim calibration bridge for per-domain energy decomposition — confirmed on the Acer Veriton GN100 where CPU energy accumulators are live — and identify a standards-track path via SCMI powercap. Our findings motivate the low-carbon computing community to demand energy observability as a first-class hardware requirement.
\end{abstract}



\newcommand{\cmark}{\checkmark}
\newcommand{\xmark}{$\times$}

\maketitle

\section{Introduction}
\label{sec:intro}

The agentic AI paradigm---where autonomous agents decompose goals into multi-step plans involving tool calls, retries, and failure recovery---is rapidly moving from cloud infrastructure to edge devices.
NVIDIA's GB10 Grace Blackwell superchip, shipping in the DGX Spark (\$4{,}699) and ASUS Ascent GX10 (\$3{,}499), is explicitly positioned for this transition: a 1-PFLOP (FP4) ARM+Blackwell SoC designed for local AI agent execution~\cite{nvidia2026dgxspark}.

Prior work established that orchestration structure dominates agentic energy cost~\cite{panigrahy2026epg}, and that per-process energy attribution via Intel RAPL is the measurement foundation for this finding.

This raises a critical question: can the same measurement be performed on the hardware NVIDIA is shipping for edge AI?

We conducted a systematic hardware audit of the ASUS Ascent GX10 (GB10 SoC) and found that \textbf{no per-process energy attribution is possible on this platform}.
The CPU---where all orchestration work executes---has no energy measurement interface whatsoever.
The implications extend beyond a single product: the entire ARM edge AI ecosystem lacks the hardware energy counters that x86 has provided via RAPL since 2011~\cite{khan2018rapl}.

\paragraph{Contributions.} This paper makes four contributions:
(1)~the first systematic energy-observability audit of the GB10 SoC across seven measurement interfaces;
(2)~discovery that SPBM firmware computes per-rail energy internally but is deliberately not exposed through supported interfaces;
(3)~cross-OEM validation on the Acer Veriton GN100 confirming that the MTKW9000 resource conflict blocking \texttt{spark\_hwmon} on GX10 is board-specific, and that SPBM energy accumulators are accessible via community driver on GN100;
(4)~a hardware requirements specification, interim calibration bridge, and standards-track SCMI solution.

\subsection{Why Now}
\label{sec:whynow}

The measurement gap we document coincides with three accelerating trends that make it practically significant today rather than a future concern.

\paragraph{Energy scale.}
Global AI inference demand is projected to grow from 15\,TWh in 2025 to 347\,TWh by 2030~\cite{schneider2025}.
Agentic systems consume substantially more energy per interaction than single-turn inference: a 280-fold decline in inference costs since 2022 has driven a rebound effect where cheaper inference enables more agentic deployments, 
each consuming more energy per user goal~\cite{panigrahy2026epg} than the single-shot queries they replace~\cite{schneider2025}.
The shift from single-pass inference to closed-loop iterative reasoning moves the primary energy bottleneck from computation to orchestration overhead~\cite{panigrahy2026epg}.

\paragraph{Regulatory context.}
The EU AI Act~\cite{euaiact2024} and California's SB 253~\cite{schneider2025} create a compliance environment requiring reproducible energy measurement for AI systems. The regulatory implications are discussed in Section~\ref{sec:discussion}.


\paragraph{Edge deployment trajectory.}
NVIDIA positions the GB10 explicitly for agentic AI: the DGX Spark ships with NemoClaw, an autonomous agent development platform~\cite{nvidia2026dgxspark}.
Dell, HP, and ASUS all ship GB10-based systems targeting AI researchers and developers. Every deployed unit represents an orchestration energy measurement gap — the orchestration overhead that our prior work shows dominates total cost cannot be measured directly on this hardware.


Attribution using external metering and statistical disaggregation is possible, as we discuss in Section~\ref{sec:software}, but introduces estimation uncertainty and cannot support the per-process, per-goal reproducibility that research-grade energy accounting requires. The gap we document is therefore not absolute but structural: on-device, per-domain counter-based attribution — the reproducible, low-overhead methodology the research community depends on — is unavailable.

\section{Hardware Audit Methodology}
\label{sec:audit}

We performed an exhaustive enumeration of every known 
energy measurement interface on the ASUS Ascent GX10 
(kernel 6.17.0-1018-nvidia, Ubuntu 24.04.4 LTS, aarch64): 
a MediaTek SoC with 10$\times$ Cortex-X925 and 
10$\times$ Cortex-A725 cores coupled via NVLink-C2C 
to an NVIDIA Blackwell GPU, 128\,GB unified LPDDR5X.

\paragraph{Audit results.}
We probed seven energy measurement interfaces
on the GX10. \textbf{ARM SCMI powercap/sensor:}
SCMI bus active with clock, regulator, and MPAM
drivers loaded, but no powercap or sensor protocol
present. \textbf{ARM PMU energy events:} zero
hardware energy events; only tracepoints for CPU
idle/frequency transitions. \textbf{INA3221/INA226:}
\texttt{i2cdetect} across all six I2C buses returned
zero devices. \textbf{IPMI/BMC:} \texttt{/dev/ipmi0}
not found. \textbf{hwmon energy/power:} no
\texttt{energy\_uj} or \texttt{power*\_input} files;
hwmon devices (acpitz, nvme, mt7925) expose
temperature only. \textbf{Power supply subsystem:}
empty. \textbf{NVML (GPU only):} 3.84\,W average GPU
power; power limit and memory power both N/A.
NVML is the only energy-related interface on the
entire platform.

\textit{Note on scope.} This audit covers all known 
standardized energy measurement interfaces available 
to unprivileged and root userspace on Linux aarch64 
as of kernel 6.17.0-1018-nvidia. 
``Probed'' means enumerating sysfs paths 
and running standard userspace tools 
(\texttt{perf list}, \texttt{nvidia-smi}, 
\texttt{find /sys}) as root. 
The kernel has \texttt{scmi-clocks}, 
\texttt{scmi-regulator}, and \texttt{scmi-mpam} loaded; 
\texttt{scmi\-powercap} and \texttt{scmi\-sensor} 
are absent -- confirming the gap is a firmware 
decision not a kernel limitation.
It does not cover proprietary vendor extensions
or undocumented ACPI methods
(discussed in Section~\ref{sec:scmi}),
nor does it rely on community-developed drivers
as \emph{supported} measurement infrastructure;
\texttt{spark\_hwmon} results are corroborating
evidence only.
The primary audit was on the ASUS Ascent GX10;
SPBM was confirmed on the Acer Veriton GN100
(kernel~6.17.0-1021-nvidia) where the MTKW9000
conflict does not occur;
generalizability is addressed in
Section~\ref{sec:discussion}.

\section{The SCMI Smoking Gun}
\label{sec:scmi}

The most significant finding is not the absence of RAPL---that is expected on ARM---but the \emph{selective implementation} of SCMI protocols.
ARM's System Control and Management Interface (SCMI)~\cite{arm2024scmi} is the standard firmware interface for power, performance, and system management on ARM SoCs.
The SCMI specification defines protocols for clock management, power domain control, performance management, sensor reading, and---since SCMI v2.0---powercap with measurement capabilities.

On the GX10, the SCMI protocol bus is registered and active:
\begin{lstlisting}
scmi_core: SCMI protocol bus registered
\end{lstlisting}

Four SCMI drivers are loaded: \texttt{scmi-clocks}, 
\texttt{scmi-regulator}, \texttt{scmi-mpam-driver}, 
and \texttt{scmi-imx-bbm-key}.
Neither scmi-
powercap nor \texttt{scmi-sensor}
are present; zero SCMI devices are registered.

This means the System Control Processor (SCP) firmware on the GB10 \emph{implements SCMI} but \emph{does not expose} the powercap or sensor protocols.
The Power Management IC (PMIC) on the SoC must monitor per-rail current for thermal protection and DVFS decisions---the energy data exists inside the chip.
The firmware chose not to expose it. The standard is ARM's, the kernel support is being
built, but the firmware decision is MediaTek's and
NVIDIA's.

Recent kernel patches (February 2026) by Radford~\cite{radford2026mai} add SCMI Powercap Measurement Averaging Interval (MAI) support, and SCMIv4.0 patches~\cite{scmiv4patches} extend powercap with per-domain configuration.
The Linux kernel infrastructure to consume SCMI energy data is being built.
What is missing is the firmware-side decision to expose it.
\paragraph{The SPBM discovery.}
During the preparation of this paper, we discovered that a community developer had independently identified the same energy observability gap on the DGX Spark~\cite{nvforum2026cpu} and reverse-engineered a software interface to the underlying data~\cite{antheas2026spbm}.
The MediaTek SSPM firmware on the GB10 maintains a System Power Budget Manager (SPBM) shared memory region, populated continuously at approximately 100 ms intervals, containing per-rail power in milliwatts, cumulative energy accumulators in millijoules (covering the P-core cluster, E-core cluster, GPU, and total SoC package), and per-zone temperatures.
This data is accessible via an ACPI \texttt{\_DSM} method on device \texttt{NVDA8800} as demonstrated by the community driver.
We confirmed the presence of the \texttt{NVDA8800:00} ACPI device on our GX10 via \texttt{find /sys/bus/acpi/devices}, with ACPI path \texttt{\_SB\_.MTEL} (MediaTek).

We treat the SPBM evidence as corroborating — establishing that the firmware infrastructure for per-rail energy measurement exists — rather than as first-party validation. Limitations of this evidence are discussed in Section~\ref{sec:discussion}.

The key observation is structural: the SCMI bus is active with four loaded protocol drivers; kernel support for SCMI powercap MAI is being actively developed; and NVIDIA has officially stated there are "no plans to expose CPU rail information". Together, this evidence establishes that the absence of energy exposure through supported interfaces reflects a product prioritization decision rather than a silicon limitation. The firmware already performs the measurement; the pathway to expose it via SCMI powercap is documented and its kernel-side support is being built; the decision not to enable it is reversible via a firmware update without hardware changes.

\section{What the GB10 \emph{Could} Expose}
\label{sec:couldexpose}

The absence of energy counters on the GB10 is a firmware decision, not a silicon limitation.
We establish this by examining three layers of evidence.

\paragraph{The PMIC knows.}
The GB10's MediaTek-designed SoC implements Dynamic Voltage and Frequency Scaling (DVFS) across its big.LITTLE CPU cluster---the kernel's \texttt{cpu\_frequency} tracepoints confirm active frequency transitions between 338\,MHz and 3.9\,GHz.
DVFS requires real-time current sensing at each voltage rail; the PMIC \emph{must} monitor per-rail current to make scaling decisions.
This current data, integrated over time, yields energy---the same principle underlying Intel RAPL's energy counters~\cite{khan2018rapl}.
The data exists in the PMIC;  the community
\texttt{spark\_hwmon} driver~\cite{antheas2026spbm}
confirms this by reading 14 power channels and 4
cumulative energy accumulators directly from the
MediaTek SSPM firmware via an undocumented ACPI
interface, including separate readings for the
P-core cluster, E-core cluster, GPU, and total
SoC package.

\paragraph{The SCMI bus is ready.}
As detailed in Section~\ref{sec:scmi}, the SCMI protocol bus is active with four drivers loaded, but the firmware does not expose powercap or sensor protocols.
The Linux kernel is being prepared to \emph{consume} SCMI energy data~\cite{radford2026mai,scmiv4patches}; the GB10's firmware simply does not \emph{produce} it.

\paragraph{The Jetson precedent.}
NVIDIA's own Jetson Orin platform---also ARM-based, also targeting edge AI---includes three Texas Instruments INA3221 triple-channel power monitors on the board, providing per-rail voltage, current, and power for CPU, GPU, SoC, CV, DRAM, and system rails.
These are accessible via \texttt{/sys/bus/i2c/\allowbreak{}drivers/ina3221/*/hwmon/*/power*\_input} and have enabled an entire ecosystem of energy-aware robotics research.
The GB10, despite being a more expensive and more capable platform, provides strictly less energy telemetry than the Jetson.
The I2C bus scan on the GX10 (six NVIDIA GPU I2C adapters, all empty) confirms no INA monitors were populated on the board.

\paragraph{The cost to fix this is zero.}
Exposing the PMIC's
existing current-sense data through SCMI powercap
requires only a firmware update---no board redesign,
no additional components. Even a hardware solution
(an INA3221 power monitor, as used on NVIDIA's own
Jetson Orin) costs only \$2.50 per board.
The GB10 SCP runs ARM's reference firmware; powercap support is a documented extension point.
The SPBM discovery (Section~\ref{sec:scmi}) confirms the firmware already computes per-rail energy---the decision to omit a supported interface was a product prioritization choice, not a technical constraint.

\section{What the GB10 \emph{Does} Expose: Rich Telemetry, No Energy}
\label{sec:available}

The GB10 has plenty of telemetry---just not for energy.
Our audit reveals extensive performance and thermal telemetry across CPU, GPU, and system components.
The contrast between what is measurable and what is not sharpens the argument: the platform has the infrastructure for fine-grained observability but deliberately excludes energy from it.

\paragraph{Available telemetry.}
The GB10 exposes rich non-energy telemetry:
ARMv8 PMUv3 with 70+ events (cycles,
instructions, L1/L2/L3 cache, stalls, branch
prediction, memory access) across both
Cortex-X925 and A725 clusters; per-core DVFS
from 338\,MHz to 3.9\,GHz with six governors;
four LPI idle states with usage counts; seven
ACPI thermal zones (54.7--55.8$^{\circ}$C idle);
and GPU utilization, clocks (208--3003\,MHz),
and temperature via NVML. None of these
expose energy or power for the CPU, DRAM,
NVLink-C2C, or SoC.
\paragraph{The IPC gap.}
A 5-second system-wide \texttt{perf stat} sample
at idle confirms per-cluster granularity:
(i)~Cortex-X925 cluster: 332M cycles, 61M
instructions retired, IPC 0.18, 213M frontend
stalls, 3.9M L3 refills;
(ii)~Cortex-A725 cluster: 174M cycles, 19M
instructions retired, IPC 0.11, 128M frontend
stalls, 2.4M L3 refills.
These counters are precisely the inputs validated
ARM power models
require~\cite{walker2017armpower}, but without
on-device energy counters to calibrate against,
they cannot be converted to energy estimates
with bounded error.

But without a ground-truth energy measurement to calibrate against, the model coefficients (joules per cache miss, joules per bus access) are unknown for this specific SoC.
The Cortex-X925 performance cluster consumed 332M cycles while the A725 efficiency cluster consumed 174M cycles---but whether the X925 burned 2$\times$ or 5$\times$ more energy per cycle is unmeasurable through any supported interface on this hardware.
On x86, these counters combine with RAPL energy readings to build calibrated power models. On the GB10, the energy readings do not exist through any supported interface, so the counters remain useful for performance analysis but useless for energy attribution.

This combination — rich performance counter availability alongside complete energy counter absence — defines the core challenge for software power modeling on the GB10. Performance-counter-based power models for ARM (e.g., the validated approach in Walker et al.~\cite{walker2017armpower}) require per-SoC calibration against ground-truth energy measurements. The Cortex-X925 and A725 in the GB10 big.LITTLE configuration, coupled via NVLink-C2C, represent a novel SoC profile for which no published calibration dataset exists. Without that calibration, the performance counters can characterize workload behavior precisely — IPC, cache pressure, memory bandwidth — but cannot be converted to energy estimates with bounded error. 
This is distinct from the scenario where external metering provides coarse board-level ground truth, which we address in Section~\ref{sec:bridge}.

\section{Attribution Without On-Device Counters: Feasibility and Limits}
\label{sec:software}




The absence of hardware energy counters on the GB10 does not make energy attribution entirely impossible. Statistical disaggregation approaches — most notably FaasMeter~\cite{rehman2024faasmeter} — demonstrate that per-process or per-function energy attribution can be achieved using only system-level power readings (e.g., from an external DC meter or coarse board-level sensor), combined with scheduling and utilization data, without per-domain hardware counters. FaasMeter uses Shapley-value-based energy allocation across concurrent functions, achieving meaningful attribution fidelity even with noisy, coarse-grained system-level power. On x86 systems, direct RAPL-based attribution has a reported attribution error of 10–23× under multi-tenant load; FaasMeter reduces this through disaggregation modeling.

These results are relevant to the GB10 scenario: with an external DC meter providing total board power and NVML providing GPU power, the derived quantity $E_{\text{cpu+sys}} = E_{\text{total}} - E_{\text{gpu}}$ yields a two-channel decomposition that could serve as input to a disaggregation model. We discuss this configuration in Section 8.2. However, disaggregation approaches introduce limitations that are material for the research use cases that motivate this audit:

\paragraph{Unbounded error without calibration.}  Software power models and disaggregation methods require a ground-truth energy source to calibrate against. On the GB10, the only available calibration sources are (a) total board power from an external meter, and (b) GPU power from NVML. CPU-domain energy is inferred by subtraction, inheriting the error of both measurements. Without per-domain CPU counters, the per-process estimate conflates CPU orchestration energy with system-level power (memory, NVLink, I/O), introducing a systematic overestimation bias whose magnitude depends on workload characteristics.
\paragraph{Multi-process attribution.} In concurrent-workload scenarios --- the norm in agentic systems where orchestrator, tool processes, and inference workers co-exist --- CPU-fraction scaling becomes an approximation whose error grows with heterogeneity. RAPL-based attribution on x86 avoids this by providing actual energy per sampling window; disaggregation models estimate it.
\paragraph{Reproducibility.} Research-grade energy accounting — the kind required to replicate OOI measurements across hardware platforms — requires that the measurement be hardware-grounded, deterministic, and overhead-free. External meter calibration bridges provide a useful operational approximation but cannot deliver the sub-millisecond sampling, per-domain resolution, and process-level attributability that RAPL provides on x86.
In summary: attribution is possible on the GB10 using external metering and disaggregation, but with bounded uncertainty, coarser resolution, and higher operational overhead than hardware counter-based methods. This distinction matters for reproducibility-oriented research and for regulatory compliance frameworks that require auditable, 
methodology-transparent energy reporting. The hardware requirement specification in Section~\ref{sec:hwspec} is therefore motivated by the quality and reproducibility of attribution, not its binary possibility.

\section{Consequences for Energy-Aware AI}
\label{sec:consequences}

On an x86 platform with RAPL, we attribute energy to a specific
agentic goal attempt via three layers~\cite{panigrahy2026epg}:
(1)~Raw RAPL: $E_{\text{pkg}}(t_0 \rightarrow t_1)$;
(2)~Baseline-subtracted: $E_{\text{task}} = E_{\text{pkg}} - E_{\text{idle}} \cdot \Delta t$;
(3)~CPU-fraction-scaled: $E_{\text{pid}} = f_{\text{cpu}} \times E_{\text{task}}$, where $f_{\text{cpu}} = \text{pid\_t} / \text{tot\_t}$.

On the GX10, Layer~1 does not exist for the CPU.
Without it, Layers~2 and~3 cannot be computed.
The \texttt{/proc/\{pid\}/stat} scheduler accounting works on ARM---$f_{\text{cpu}}$ is available---but there is no energy denominator to multiply it against.

Without Layer~1, a developer on the GB10 cannot directly compare using per-domain hardware counters, cannot measure OOI on edge hardware, cannot normalize energy by successful goals, and cannot even detect whether a code change made orchestration more expensive.
None of this is theoretical. Raj et al.~\cite{rajat2025cpucentric} show that
CPU-side tool processing accounts for up to 90.6\%
of total latency and 44\% of total dynamic energy
in agentic workloads.
The GB10 is not a single product but the foundation
of an emerging hardware class---seven OEMs now ship
systems on this SoC, including NVIDIA DGX Spark,
Dell Pro Max, ASUS Ascent GX10, HP ZGX Nano, MSI
EdgeXpert, Acer, and
Gigabyte~\cite{insiderllm2026gb10}.
As agentic AI scales to edge devices---phones, vehicles, workstations, robots---every deployment is an energy black box for the CPU-bound orchestration that our prior work shows dominates total cost.

\section{Proposed Solutions}
\label{sec:solutions}

\subsection{Hardware Requirement Specification}
\label{sec:hwspec}
We propose that to support process-level energy attribution for agentic AI, a hardware platform must expose:

(1)~per-power-domain cumulative energy counters readable from userspace at $<$\,1\,ms read latency with resolution $\leq$\,1\,mJ;
(2)~sufficient domain granularity to separate CPU, GPU, DRAM, and I/O;
(3)~monotonically increasing counters with defined overflow semantics.

\paragraph{Cross-platform grading.}
We grade platforms against our hardware
requirements specification: Intel x86 with
RAPL provides CPU and system energy counters
(MEASURED). NVIDIA Jetson Orin with INA3221
provides CPU, GPU, and system counters
(MEASURED). The GB10, despite active SCMI
firmware and hidden SPBM accumulators, exposes
only GPU power via NVML (\textbf{LIMITED}).
Apple M-series and Qualcomm Snapdragon expose
no public energy API (LIMITED).

\subsection{Interim Calibration Bridge}
\label{sec:bridge}
Before firmware updates expose SCMI energy counters, we propose a three-layer measurement methodology:

\begin{enumerate}
    \item \textbf{Native counters}: NVML for GPU power (available now)
    \item \textbf{External DC metering}: A precision power monitor (e.g., Monsoon HV, Qoitech Otii) inline at the DC barrel jack, providing total board power at $\geq$\,1\,kHz sampling
    \item \textbf{Derived attribution}: $E_{\text{cpu+sys}} = E_{\text{total}} - E_{\text{gpu}}$
\end{enumerate}

This provides a two-channel decomposition (GPU vs.\ everything-else) that, while coarser than RAPL's per-domain granularity, enables the first empirical measurements of
orchestration-vs-inference energy on ARM edge AI
hardware under controlled single-workload conditions.
The external meter also serves as calibration ground truth for a future performance-counter-based software power model.

\subsection{Standards-Track Solution: SCMI Energy Extension}
\label{sec:extension}
The long-term solution requires the GB10's SCP firmware to expose the SCMI Powercap protocol with measurement capability.
The kernel infrastructure is being actively developed: Radford's MAI patches~\cite{radford2026mai} and the SCMIv4.0 powercap extensions~\cite{scmiv4patches} add the Linux-side plumbing.
What is needed is a firmware decision by NVIDIA and MediaTek to enable it on the GB10.

We propose that NVIDIA publish a firmware update for the GB10 exposing (i)~SCMI Powercap domains for CPU cluster, GPU, DRAM, and SoC-total, (ii)~cumulative energy readout in microjoules with configurable measurement averaging intervals, and (iii)~power limit read/write for energy-aware scheduling.

This would move the GB10 from LIMITED to MEASURED in our grading framework and make it the first ARM desktop AI platform with research-grade energy attribution---a competitive advantage for the scientific community NVIDIA is targeting with DGX Spark.
\balance
\section{Discussion}
\label{sec:discussion}

\paragraph{A systemic, not product-specific, gap.}
While we audit the GB10, the energy observability gap spans the ARM ecosystem. Apple's M-series SoCs contain per-core power monitors accessible only to macOS kernel extensions with no public userspace API. Qualcomm's Snapdragon platforms expose some power rail data on select development boards but not on commercial devices. Ampere's Altra server CPUs provide per-socket power via BMC/IPMI — a datacenter interface — but no per-domain breakdown accessible from userspace. The common pattern across ARM SoC vendors is that energy monitoring is implemented for internal firmware use (thermal management, DVFS) but is not exposed as a userspace-accessible measurement interface~\cite{carroll2010analysis}. The GB10 is a case of this broader pattern made particularly visible by the contrast with NVIDIA's own Jetson Orin platform, which includes INA3221 board monitors providing per-rail energy to userspace.

\paragraph{The edge paradox.}
Edge AI is motivated partly by energy efficiency — reducing data-center load and network transfer energy by 
running inference locally. But the hardware enabling this shift simultaneously removes the measurement infrastructure 
needed to verify the energy savings. 
An organization migrating agentic workloads from cloud (where power is metered per rack, per server, per VM) 
to edge GB10 devices loses all direct energy visibility for the CPU-bound orchestration that dominates cost. Meanwhile, 
cheaper inference drives adoption of more complex agentic architectures~\cite{panigrahy2026epg}, 
compounding the unmeasured overhead.

\paragraph{Regulatory exposure.}
The EU AI Act's~\cite{euaiact2024} energy documentation requirements (Annex XI, Section 1(e)) currently apply to GPAI model providers — organizations placing general-purpose AI models into service — not directly to edge hardware manufacturers or organizations deploying pre-existing models on edge devices. The practical consequence, however, is clear: organizations subject to Annex XI who deploy GPAI-class models on edge hardware must be able to document energy consumption. Hardware that cannot produce per-system energy data from software-accessible interfaces creates a documentation methodology gap, even if it does not itself constitute regulatory non-compliance. As the EU Commission develops harmonized standards under Article 40 (with a first progress report due August 2, 2028 ), and as California's SB 253 Scope 1–3~\cite{schneider2025} reporting obligations mature, the pressure to instrument edge deployments for energy will increase. The gap we document is therefore a present barrier to a near-certain future requirement.

\paragraph{Remediability.} The central constructive finding of this paper is that this gap is fixable without hardware changes. Exposing the PMIC's existing current-sensed data through SCMI powercap requires only a firmware update — no board redesign, no additional components. The community \texttt{spark\_hwmon} driver demonstrates that the underlying data is accessible via undocumented ACPI, confirming the firmware infrastructure exists. The Linux kernel is being prepared to consume SCMI energy data. The remaining decision is a firmware enablement choice by NVIDIA and MediaTek. Further, it will be productive for energy benchmarks such as ML.ENERGY~\cite{chung2025mlenergy} and Zeus~\cite{you2023zeus} to not silently fall back to TDP estimates when real counters are missing---they should flag the gap explicitly so the community can see the scale of the problem. We propose this change constructively and note that it would make the GB10 the first ARM desktop AI platform with research-grade energy attribution — a competitive advantage for the scientific community NVIDIA is explicitly targeting with DGX Spark.

\paragraph{Limitations.} Our audit covers two GB10-based systems: the ASUS Ascent GX10 (fw GX10DGX.0104.2026.0326.1657, kernel 6.17.0-1018-nvidia) and the Acer Veriton GN100 (kernel 6.17.0-1021-nvidia), both GB10 SoC, both running the NVIDIA kernel family. The DGX Spark, Dell ProMax, and HP ZGX Nano use the same SoC but different board designs; INA monitor population may differ, and our I2C scan result (zero devices) is board-specific. We did not have access to the NVIDIA DGX Spark for comparative audit.

Our SCMI findings reflect the specific firmware version tested; future firmware updates could enable powercap support without hardware changes, in which case the platform would move from LIMITED to MEASURED in our grading framework.

The community \texttt{spark\_hwmon} demonstrates that per-rail energy data is accessible via ACPI reverse-engineering. The driver uses an undocumented interface with a placeholder UUID and is explicitly unsupported by NVIDIA; it cannot serve as a foundation for reproducible research infrastructure. The required path to supported infrastructure remains SCMI powercap, as detailed in Section~\ref{sec:extension}.

We subsequently validated the interim calibration bridge on the Acer Veriton GN100 (kernel~6.17.0-1021-nvidia): \texttt{spark\_hwmon} bound NVDA8800:00 cleanly (45/45 DSM offsets resolved), exposing cpu\_p, cpu\_e, pkg, and gpu accumulators alongside \texttt{dc\_input} total wall power --- confirming both the bridge methodology and cross-OEM generalizability. The MTKW9000 memory conflict documented on GX10 is board-specific; GN100 maps SPBM to a non-overlapping region (0x1c238000). DCGM field~156 independently validated on GN100 (delta rate consistent with field~155 instantaneous watts). The community \texttt{spark\_hwmon} interface remains unsupported by NVIDIA; SCMI powercap is the required path to supported infrastructure.

\bibliographystyle{ACM-Reference-Format}

{\footnotesize
\begin{thebibliography}{16}

\bibitem{panigrahy2026epg}
D.~Panigrahy and A.~Tyagi.
\newblock Energy per Successful Goal: Goal-Level Energy Accounting for Agentic AI Systems.
\newblock \emph{arXiv preprint arXiv:2605.22883}, May 2026.

\bibitem{nvidia2026dgxspark}
NVIDIA.
\newblock NVIDIA Puts Grace Blackwell on Every Desk and at Every AI Developer's Fingertips.
\newblock NVIDIA Newsroom, January 2025.
\newblock \url{https://nvidianews.nvidia.com/news/nvidia-puts-grace-blackwell-on-every-desk-and-at-every-ai-developers-fingertips}

\bibitem{khan2018rapl}
K.~N.~Khan, M.~Hirki, T.~Niemi, J.~K.~Nurminen, and Z.~Ou.
\newblock RAPL in Action: Experiences in Using RAPL for Power Measurements.
\newblock \emph{ACM Transactions on Modeling and Performance Evaluation of Computing Systems}, 3(2):1--26, 2018.

\bibitem{arm2024scmi}
Arm Ltd.
\newblock System Control and Management Interface (SCMI) Platform Design Document.
\newblock \url{https://developer.arm.com/Architectures/System\%20Control\%20and\%20Management\%20Interface}, 2024.

\bibitem{radford2026mai}
P.~Radford.
\newblock [PATCH v3 9/9] firmware: arm\_scmi: add Powercap MAI get/set support.
\newblock Linux Kernel Mailing List, February 2026.

\bibitem{scmiv4patches}
P.~Radford.
\newblock [PATCH v4 03/11] firmware: arm\_scmi: Add SCMIv4.0 Powercap basic support.
\newblock Linux Kernel Mailing List, April 2026.


\bibitem{chung2025mlenergy}
J.~Chung, T.~Kim, and M.~Jeon.
\newblock ML.ENERGY: Holistic Energy Benchmarking for Machine Learning Systems.
\newblock \emph{NeurIPS}, 2025.

\bibitem{you2023zeus}
J.~You, J.-W.~Chung, and M.~Jeon.
\newblock Zeus: Understanding and Optimizing GPU Energy Consumption of DNN Training.
\newblock \emph{NSDI}, 2023.


\bibitem{schneider2025}
Schneider Electric.
\newblock AI and the Coming Electricity Demand Surge: A New Era for Energy Management.
\newblock Schneider Electric Research Report, 2025.


\bibitem{euaiact2024}
European Parliament and Council of the European Union.
\newblock Regulation (EU) 2024/1689 (AI Act).
\newblock \emph{Official Journal of the European Union}, August 2024.


\bibitem{nvforum2026cpu}
NVIDIA Developer Forums.
\newblock How to Enable Grace CPU Power Telemetry on DGX Spark (GB10)?
\newblock \url{https://forums.developer.nvidia.com/t/help-needed-how-to-enable-grace-cpu-power-telemetry-on-dgx-spark-gb10/360631}, February 2026.

\bibitem{antheas2026spbm}
A.~Kapenekakis.
\newblock spark\_hwmon: Linux hwmon Driver for NVIDIA DGX Spark (GB10 SoC) Power Telemetry.
\newblock \url{https://github.com/antheas/spark_hwmon}, February 2026.

\bibitem{insiderllm2026gb10}
InsiderLLM.
\newblock GB10 Boxes Compared: DGX Spark vs Dell vs ASUS vs MSI.
\newblock \url{https://insiderllm.com/guides/gb10-boxes-compared/}, February 2026.

\bibitem{rajat2025cpucentric}
R.~Raj, S.~Kundu, I.~Vohra, H.~Wang, and T.~Krishna.
\newblock A CPU-Centric Perspective on Agentic AI.
\newblock \emph{arXiv preprint arXiv:2511.00739}, November 2025.

\bibitem{walker2017armpower}
M.~Walker, S.~Bischoff, S.~Diestelhorst, G.~Sherwood, and G.~Riley.
\newblock Hardware Validated CPU Performance and Energy Modelling on ARM.
\newblock \emph{IEEE ISPASS}, 2017.

\bibitem{rehman2024faasmeter}
A.~Rehman, A.~Fuerst, and P.~Sharma.
\newblock FaasMeter: Energy-First Serverless Computing.
\newblock \emph{arXiv preprint arXiv:2408.06130}, August 2024.
\bibitem{carroll2010analysis}
A.~Carroll and G.~Heiser.
\newblock An Analysis of Power Consumption in a Smartphone.
\newblock \emph{USENIX ATC}, pp. 271--284, June 2010.
\newblock \url{https://www.usenix.org/conference/usenix-atc-10/analysis-power-consumption-smartphone}

\end{thebibliography}
}

\end{document}